\documentclass[a4paper,twoside]{article}

\usepackage{epsfig}
\usepackage{subcaption}
\usepackage{calc}
\usepackage{amssymb}
\usepackage{amstext}
\usepackage{amsmath}
\usepackage{amsthm}
\usepackage{multicol}
\usepackage{pslatex}
\usepackage{apalike}

\usepackage{csquotes}
\usepackage{graphicx}
\usepackage{xspace}
\usepackage{cleveref}
\usepackage{booktabs}

\usepackage{SCITEPRESS}     

\newcommand*{\wrt}{w.r.t.\@\xspace}
\newcommand*{\eg}{\emph{e.g.,}\@\xspace}
\newcommand*{\ie}{\emph{i.e.,}\@\xspace}

\begin{document}

    \title{Self-Supervised Learning from Semantically Imprecise Data}

    \author{\authorname{Clemens-Alexander Brust\sup{1}\orcidAuthor{0000-0001-5419-1998}, Björn Barz\sup{2}\orcidAuthor{0000-0003-1019-9538} and Joachim Denzler\sup{1,2}\orcidAuthor{0000-0002-3193-3300}}
    \affiliation{\sup{1}DLR Institute of Data Science, Jena, Germany}
    \affiliation{\sup{2}Friedrich Schiller University Jena, Jena, Germany}
    \email{\{clemens-alexander.brust, bjoern.barz, joachim.denzler\}@uni-jena.de}
    }

    \keywords{Imprecise Data, Self-Supervised Learning, Pseudo-Labels}

    \abstract{%
        Learning from imprecise labels such as \enquote{animal} or \enquote{bird}, but making precise predictions
        like \enquote{snow bunting} at inference time is an important capability for any classifier when expertly labeled training data is scarce.
        Contributions by volunteers or results of web crawling lack precision in this manner, but are still valuable.
        And crucially, these weakly labeled examples are available in larger quantities for lower cost than high-quality bespoke training data.
        CHILLAX, a recently proposed method to tackle this task, leverages a hierarchical classifier to learn from imprecise labels.
        However, it has two major limitations.
        First, it does not learn from examples labeled as the root of the hierarchy, \eg{} \enquote{object}.
        Second, an extrapolation of annotations to precise labels is only performed at test time,
        where confident extrapolations could be already used as training data.
        In this work, we extend CHILLAX with a self-supervised scheme using constrained semantic extrapolation to generate pseudo-labels.
        This addresses the second concern, which in turn solves the first problem, enabling an even weaker supervision requirement than CHILLAX.
        We evaluate our approach empirically, showing that our method allows for a consistent accuracy improvement of $0.84$ to $1.19$ percent points over
        CHILLAX and is suitable as a drop-in replacement without any negative consequences such as longer training times.
    }

    \onecolumn \maketitle \normalsize \setcounter{footnote}{0} \vfill
    \section{INTRODUCTION}
    High-quality training data labeled by domain experts is an essential ingredient for a
    successful application of contemporary deep learning methods.
    However, such data is not always available or affordable in sufficient quantities.
    And while there exist a number of effective strategies to maximize sample efficiency such as
    data augmentation, transfer learning, or active learning, there are also limits to the information
    that can be extracted from a small dataset.

    If larger quantities of training data are required, a compromise \wrt{} the quality has to be made, \ie{} by allowing noisy labels.
    Such data is available at lower cost and greater quantity, \eg{} by employing volunteer labelers instead of experts
    or crawling the web for training examples.

    In this case, lower quality means that the labels are noisy \wrt{} to two aspects: accuracy and precision.
    Inaccuracy means that labels are simply incorrect, \ie{} confused with other classes.
    Imprecise labels are correct, but carry less semantic information in terms of depth in a class hierarchy (see \cref{fig:classhierarchy}), \eg{} \enquote{animal} vs. \enquote{bird}.
    In \cite{Brust2020ELC}, this weakly supervised task is formally defined as \enquote{learning from imprecise data}.
    It is defined such that the training data can contain imprecise labels,
    but predictions must always be as precise as possible, \ie{} leaf nodes of the hierarchy.
    At test time, labels are said to be \emph{extrapolated} (from imprecise to precise) by their method CHILLAX, which we briefly explain in \cref{sec:sidexplained}.

    While their method can perform the task reliably, it has two main limitations.
    The disadvantages come from the underlying probabilistic hierarchical classifier \cite{brust_integrating_2018}.
    The classifier is modified to perform the extrapolation at test time and to accept imprecise labels during training.
    However, it cannot learn from examples that are labeled at root of the hierarchy, \eg{} as \enquote{object},
    even though it is clearly capable of the necessary extrapolation at test time.
    And while it can learn from inner node examples, it does not take any advantage of training time extrapolation for such examples either.
    
    Our main contribution in this work is a self-supervised approach to learning from imprecise data based on pseudo-labels.
    To avoid learning mispredictions and feedback loops, we describe several strategies constraining the extrapolation from imprecise labels to more precise pseudo-labels.
    These strategies are based on prediction confidence scores and on the structure of the hierarchy.
    We also propose methods that are less sensitive to changes in confidence score distributions over time, which we call adaptive.

    The experimental evaluation concerns two areas.
    First, we assess the potential and limits of extrapolation techniques by examining a best case scenario free of feedback loops.
    We then evaluate the performance of our methods against CHILLAX and observe the effects of a large range of parameters.
    The results show consistent improvements of $0.84$ to $1.19$ percent points in accuracy.
    All experiments are performed on the North American Birds dataset \cite{van_horn_building_2015} for direct comparison to \cite{Brust2020ELC}.

    \section{RELATED WORK}
    The task of learning from \emph{semantically imprecise} labels is proposed in
    \cite{Brust2020ELC}, where a class hierarchy is used to formally define it.
    It is then tackled using a modified hierarchical classifier \cite{brust_integrating_2018}.
    However, if an image is labeled at the root of the hierarchy, this classifier cannot leverage it.
    Instead, the image is ignored entirely.
    This property is a result of the closed world assumption, which most hierarchical classifiers make \cite{silla_survey_2011}.
    In this work, we resolve this deficiency by extending the method in \cite{Brust2020ELC} with a self-supervision scheme.

    Labels can be imprecise in other respects, \eg{} missing labels in multi-label classification \cite{abassi_imprecise_2020},
    or a set of labels where only one is expected \cite{ambroise_learning_2001}.
    An important problem of semantically imprecise labels is that the classes are no longer mutually exclusive.
    The associated consequences are discussed in detail in \cite{mcauley2013optimization},
    altough this work does not consider label extrapolation. Instead, it allows imprecise predictions.
    In \cite{deng2012hedging}, the authors explicitly mention the trade-off between accuracy and precision (specificity in their terms) and propose an algorithm that can reduce the precision of predictions such that a certain accuracy is guaranteed.
    This task is the opposite of ours, where the precision of labels is reduced, but the predictions are as precise as possible.

    The term \emph{self-supervised learning} has different meanings depending on the specific field.
    It is commonly used in unsupervised tasks such as visual representation learning
    \cite{Kolesnikov_2019_CVPR}.
    In this setting, the supervision that makes training a deep neural network possible comes from solving auxilliary or \enquote{pretext} tasks like predicting the previously applied rotation of an image.
    Real applications should benefit from the representations learned on the auxilliary tasks because unlabeled images are ubiquitous.
    
    Another common interpretation of self-supervision is also known as \emph{pseudo-labeling}, where confident predictions of a model are used as training data
    \cite{lee_pseudo-label_2013,sohn_fixmatch_2020,wang_cost-effective_2016}.
    We use this definition in our work and focus on interpreting the confidence scores at each level and node in the class hierarchy correctly to maximize the reliability of our pseudo-labels.
    Auxilliary tasks and pseudo-labeling approaches can also be combined \cite{Zhai_2019_ICCV}.

    \section{SELF-SUPERVISED METHOD}
    In this section, we describe our proposed methods and their theoretical background.
    We first review the concept of semantically imprecise data and the existing method to learn from this data.
    Then, we introduce our self-supervised approaches.

    \subsection{Semantically Imprecise Data}\label{sec:sidexplained}
    \begin{figure}
        \centering{}
        \includegraphics[scale=1]{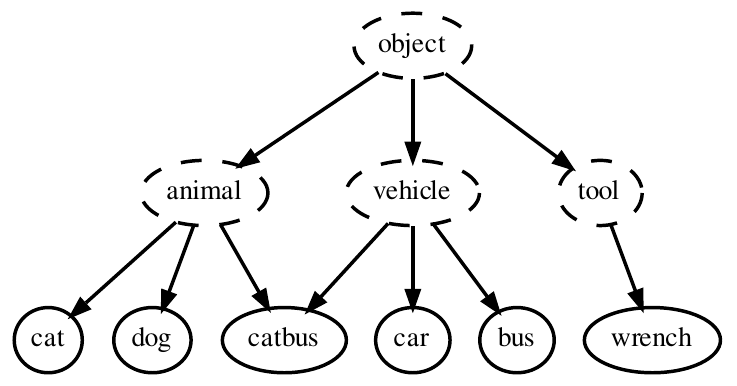}
        \caption{A class hierarchy with semantically imprecise (dashed) and precise (bold) classes.}
        \label{fig:classhierarchy}
    \end{figure}


    Given a class hierarchy, \eg{} \cref{fig:classhierarchy}, we can distinguish between semantically precise and
    imprecise labels. Precise labels are leaf nodes in the hierarchy, while imprecise labels consist of inner nodes and the root.
    A useful analogy is the number of digits of a measurement.
    Like the depth in a class hierarchy, a high number does not guarantee an accurate measurement -- only a precise one.

    The term \emph{imprecise data} is used as a shorthand to describe training data that can contain examples with semantically imprecise labels
    \cite{Brust2020ELC}.
    This relaxation does not apply to the predictions, \ie{} we still expect them to be as precise as possible.
    If the goal is to allow more training data to be used in order to improve an existing application, the predictions should remain unchanged.
    An extrapolation from imprecise training data to precise output is taking place.

    In \cite{Brust2020ELC}, the CHILLAX method based on \cite{brust_integrating_2018} is proposed.
    Instead of predicting probabilities for each class given some input, a deep neural network predicts the probability of a class \(s\) being present
    (the event \(Y_s^+\)),
    conditioned on
    both the input image $X$ and the presence \(Y_{S'}^+\) of \emph{any} parent class (cf.~\cite[eq.~(5)]{brust_integrating_2018}):
    \begin{equation}
        P(Y_s^+|X) = \underbrace{P(Y_s^+|X, Y_{S'}^+)}_{\text{DNN output}} \cdot{} P(Y_{S'}^+|X)
    \end{equation}

    This equation is evaluated recursively to obtain the final unconditional probabilities for all \enquote{allowed} classes $s$,
    \ie{} leaf nodes in the hierarchy.
    The final prediction is the leaf node with the highest probability.
    Restricting the possible predictions is necessary, as leaf nodes cannot have a higher probability than associated inner nodes owing to the multiplication of more factors that are $\leq{} 1$.
    Crucially, the recursion ends at the root where the probability is always one, encoding the closed-world assumption.
    Hence, training examples that are only labeled at the root have no effect on the classifier in terms of loss and thus no value.

    \subsection{Self-Supervised Approach}
    We propose to make use of effectively unlabeled data at the root of the class hierarchy,
    and increase the value of other imprecise examples, by extrapolating their labels
    to pseudo-labels during training.
    The classifier's own predictions are used to generate the pseudo-labels.
    Results in \cite{Brust2020ELC} show that extrapolation at test time is reliable and we examine this in more detail in \cref{sec:limitsofselfsupervision}.

    However, not all predictions are correct, and using incorrect labels for training can be worse than ignoring the respective example altogether (although controlled amounts of incorrect labels can also be benefical, see \cite{xie2016disturblabel}).
    But because we work in a hierarchical setting, there is a middle ground between ignoring and using predictions as ground truth.
    Instead of leaf nodes, labels can also be extrapolated to internal nodes where there is less potential for confusion.
    In the following, we propose several methods to determine the appropriate level of extrapolation.

    \subsubsection{Non-Adaptive Methods}\label{sec:nonadaptivemethods}
    While we require that predictions are precise, the same is not true for training data, and crucially, also not for pseudo-labels.
    We can potentially extrapolate an imprecise label (the source) to a slightly more precise label (the target), while both are inner nodes of the hierarchy.
    Our main selection criteria for extrapolation targets are the unconditional probabilities \(P(Y_s^+|X)\), which we compute for all classes in the hierarchy.
    Using predictions for all nodes is clearly not necessary, if the label is sufficiently precise, because that would ignore the label completely.
    Instead, we replace all predicted probabilities with $1$ or $0$ where appropriate, which also improves the predictions of related nodes through the recursion.
    This process ensures that the extrapolation target never \enquote{disagrees} with the source.
    Semantically, the source always subsumes the target.

    However, because these unconditional probabilities are always higher for nodes closer to the root, we cannot simply choose the most confident prediction.
    Instead, we consider the following three approaches that work by building a list of candidates (initially always all classes subsumed by the source class), then excluding some, and finally sorting the remainder to make a selection.
    
    \begin{enumerate}
        \item [(a)] \emph{Leaf Node} Extrapolate any label to the most probable leaf node.
        Effectively, all predictions are used as training data without any further consideration.
        This strategy has the highest potential for improvement, but also for inaccuracy.
        There are also no parameters to tune.

        \item [(b)] \emph{k Steps Down} Limit extrapolation to exactly $k$ steps \enquote{down} the hierarchy from the label,
        \ie{} in the more precise direction.
        Leaf nodes that are less than $k$ steps down are also allowed because there might not be any possible extrapolations otherwise.
        The label is selected based on the highest predicted unconditional probability%
        \footnote{\label{ftn:sortingnoise}Note that we add gaussian noise with $\sigma=0.0001$
        to the probabilities before sorting to make it intentionally unstable.
        This is necessary because many predicted probabilities are exactly $0.5$ during initial training as a result of intially zero weights and a sigmoid activation function.
        If the sorting were stable, the resulting order would be biased by outside factors such as memory layout.}%
        . A threshold on the probabilities can be applied optionally to further exclude unconfident predictions.

        \item [(c)] \emph{Fixed Threshold} Extrapolate only to labels with a predicted unconditional probability
        greater than or equal to a given, fixed threshold, \eg{} $0.8$. We sort all candidates by this probability%
        \footnotemark[1].
        Afterwards, a stable sort of the candidate labels by information content (IC, a measure of semantic precision)
        is performed, such that the candidate with the highest overall IC is first.
        Thus, if two candidates have the same IC, we select the candidate with higher predicted probability.
        We use the formulation of IC given in \cite[55, eq.\,3.8]{harispe_semantic_2015}.
    \end{enumerate}

    If no candidate remains in either (b) or (c), the extrapolation source is used as the target.
    The approaches (a-c) are \emph{stateless} and require no information other than the extrapolation source, class hierarchy, and predicted probablities from the deep neural network.
    
    \subsubsection{Adaptive Methods}\label{sec:adaptivemethods}
    A fixed treshold as described in the previous section is intuitive and easy to implement.
    However, it has major disadvantages.
    First, it has to be fine-tuned carefully. 
    Second, a threshold that is optimal in one training step may not be optimal for the next
    because of the continually increasing confidence during training.
    
    While methods (a) and (b) do not rely on a threshold and thus don't suffer from
    these two effects, those methods are often outperformed by a fixed threshold.
    The experiments in \cref{sec:expnonadaptive} show that constraining possible extrapolations is critical, and that a fixed structural criterion as in (b) is not always sufficient.

    Instead of relying on probabilities directly, we propose considering the difference in IC
    of the label before and after extrapolation, IC gain in short.
    IC is a more meaningful measure than hierarchical distance because it also takes global and local properties, \eg{} fan-out, of the graph into account.
    However, we still want to consider the predicted probabilities as an indicator of confidence in certain classes.
    
    To achieve both goals at the same time, we propose two \emph{adaptive} methods.
    They are adaptive in the sense that the specific selection criteria vary based on the source label:
    \begin{enumerate}
        \item[(d)] \emph{Adaptive Threshold} We maintain a moving average of the last $64$ IC gains $\bar{h}$ and strive for a target IC gain $h^*$.
        This average is used to calculate a probability threshold $\theta$ for each time step $t$ (representing weight updates or minibatches) by applying a simple update rule:
        \begin{equation}
            \theta^{(t)} = \theta^{(t-1)} + \bar{h} - h^*.
        \end{equation}
        The threshold is bounded by $[0.55,1.0]$ and initially set to the lower bound. We then apply the sorting algorithm in (c) to perform the actual extrapolation using the current value of \(\theta\).

        \item[(e)] \emph{IC Range} An interval of allowed IC differences is defined, \eg{} $[0.1,0.3]$, with the target IC gain in the middle.
        We use the interval to preselect extrapolation candidates.
        Depending on the extrapolation source, the IC difference range allows varying hierarchical distances to the target,
        as opposed to the fixed criterion in (b).
        Thus, we still consider this approach adaptive even though the parameters stay constant throughout training.
        A final selection on the preselected candidates is made using (c) with a threshold of $0.55$ to remove further spurious predictions.
    \end{enumerate}

    Both proposed methods make use of the predicted probabilities through application of the algorithm in (c), but rely mainly on its sorting and less on the threshold.
    Furthermore, (d) is \emph{stateful} because the value of \(\theta\) needs to be preserved across iterations, which may be a disadvantage from an implementation perspective.

    \section{EXPERIMENTS}
    This section contains the experimental evaluation of our methods.
    We start with a study that determines an upper limit for any gains from self-supervision by examining the pseudo-labels generated by our methods with hierarchical error measures.
    Then, we test the efficacy of our approaches in a benchmark setting as well as their sensitivity to parameters.
    This detailed evaluation is seperated into non-adaptive and adaptive approaches as described in \cref{sec:nonadaptivemethods} and \cref{sec:adaptivemethods}, respectively.

    \subsection{Setup}
    We evaluate the effects of adding our proposed self-supervised schemes to the CHILLAX method \cite{Brust2020ELC}.
    A conventional one-hot softmax classifier is added as a baseline for comparison.
    Our experimental setup generally matches that of \cite{Brust2020ELC},
    except for adjustments to the learning rate ($\eta$) and the $\ell_2$ regularization coefficient $\beta$.
    We use $\eta=0.0044,\beta=5 \cdot 10^{-5}$ and $\eta=0.003,\beta=1.58114 \cdot 10^{-5}$ for CHILLAX and 
    the one-hot baseline, respectively.

    We report results on the North American Birds (NABirds) dataset \cite{van_horn_building_2015} for comparison to the original CHILLAX method.
    This fine-grained classification dataset consists of approx. \(48500\) images of \(555\) species of birds.
    Crucially, it is also equipped with a class hierarchy.
    The training portion of the NABirds dataset is modified in different ways according to a selection of noise models from \cite{Brust2020ELC}:
    \begin{enumerate}
    \item[(i)] No noise ($100\%$ precise labels),
    \item[(ii)] Relabeling to parent with $p=0.99$ \cite{deng_large-scale_2014} ($1\%$ precise labels),
    \item[(iii)] Geometric distribution with $q=0.5$ ($9.6\%$ precise labels),
    \item[(iv)] Poisson distribution with $\lambda=1$ ($4.8\%$ precise labels),
    \item[(v)] Poisson distribution with $\lambda=2$ ($22.7\%$ precise labels).
    \end{enumerate}
    These noise models cover a wide range of scenarios from expertly labeled high-quality data (i) to web crawling (iii) and volunteer labelers (iv,v).
    We also include the protocol (ii) which is proposed in \cite{deng_large-scale_2014}.
    The models are distributions over depths in the class hierarchy, which are realized by replacing the original precise labels with parent classes to match the distribution.
    Note that our experiments are limited to label noise in terms of imprecision.
    Inaccuracy is beyond the scope of this work, but the extent of CHILLAX's relative robustness against inaccurate labels is demonstrated in \cite{Brust2020ELC}.

    Source code will be made available publicly upon formal publication.

    \subsection{Limits of self-supervision}\label{sec:limitsofselfsupervision}
    \begin{table*}
        \caption{Hierarchical F1 (\%) on NABirds validation set. Comparison between ground truth and extrapolated noisy validation labels after learning
        noisy training data. (\textnormal{i}) is not included as all results are $100\%$.}
        \label{tbl:limitsofselfsupervision}
        \centering{}
        \begin{tabular}{lrrrrr}
            \toprule
            Method / Noise & (ii) & (iii) & (iv) & (v) \\
            \midrule
            Baseline (No Extrapolation)           &           87.49 &           59.79 &            61.73 &                78.06 \\
            $1$ Step Down        &           94.74 &           73.97 &            74.81 &                87.27 \\
            $2$ Steps Down       &           94.74 &           79.29 &            79.03 &                91.31 \\
            $3$ Steps Down       &           94.72 &           80.88 &            79.29 &                92.91 \\
            Fixed Threshold $0.55$     &           90.99 &           \textbf{82.87} &            \textbf{81.37} &                \textbf{93.66} \\
            Fixed Threshold $0.8$      &           90.72 &           81.19 &            80.03 &                92.99 \\
            Leaf Node          &           \textbf{94.75} &           81.15 &            79.12 &                93.20 \\
            \bottomrule
        \end{tabular}
    \end{table*}

    \begin{table*}
        \caption{Accuracy (\%) on NABirds validation set. Comparison between ground truth and extrapolated noisy validation labels after learning
        noisy training data. (\textnormal{i}) is not included as all results are $100\%$.}
        \label{tbl:limitsofselfsupervisionaccuracy}
        \centering{}
        \begin{tabular}{lrrrrr}
            \toprule
            Method / Noise & (ii) & (iii) & (iv) & (v) \\
            \midrule
            Leaf Node          &     76.99 & 58.52 & 51.03 & 83.81 \\
            Leaf Node From Root & 63.09 & 49.55 & 43.36 & 70.99\\
            \bottomrule
        \end{tabular}
    \end{table*}

    Self-supervised learning with pseudo-labels is subject to a feedback loop
    when high-confidence predictions are used as training data.
    The scores, \eg{}:

    \begin{equation*}
        (0.02,0.73,0.11,\ldots{})\,,
    \end{equation*}

    {\setlength\parindent{0pt} are extrapolated to a maximally confident label:}
    
    \begin{equation*}
        (0.0, 1.0, 0.0,\ldots{})\,,
    \end{equation*}

    {\setlength\parindent{0pt} in addition to a potential extrapolation from one class to another.}
    Subsequent predictions of the same image are likely to be even more confident,
    thereby increasing the chance of selecting the same target class for training again.
    This feedback loop can lead to overfitting, learning of potentially false pseudo-labels,
    and overrepresentation of a subset of training data.
    While not all methods proposed in this work are subject to this effect, 
    it is important to investigate the potential and limits of extrapolated pseudo-labels in a controlled manner such that feedback effects do not influence the evaluation.

    To this end, we train one CHILLAX classifier on a noisy NABirds training set for each of the noise models (i)-(v).
    We then make predictions for the unseen NABirds validation data, whose labels we also modify using (i)-(v) and use as extrapolation sources.
    
    \Cref{tbl:limitsofselfsupervision} shows the hierarchical F1 (hF1) score between the noise-free ground
    truth and the extrapolated noisy label from our non-adaptive methods as well as a \enquote{do nothing} baseline.
    In terms of hF1, all methods outperform the baseline. This is not surprising in and of itself.
    Since no method will select an extrapolation target that disagrees with the imprecise source, it is impossible for them to perform worse than the baseline, at least in terms of hierarchical recall.
    Still, the level of outperformance is substantial in all cases and hierarchical precision could still decrease.

    Observing noise model (ii), the theoretically largest possible improvement over the baseline is $12.51$ percent points (pp), if all labels are correctly extrapolated to their precise origin.
    The actual improvements realized by our approaches range from $3.32$ pp to $7.26$ pp over the baseline.
    In contrast, we observe the largest overall improvement of $28.08$ pp over the baseline out of a theoretically possible $40.21$ pp in setting (iii).
    This is the noisest model, \ie{} the one with the lowest expected depth in the class hierarchy.
    
    \textbf{In all cases, the best evaluated approaches can fill more than half of the performance gap from the baseline to the precise labels in terms of hF1.}

    We can also evaluate the extrapolated labels in terms of classification accuracy, but this is only possible for the \enquote{leaf node} approach.
    The other approaches produce imprecise labels as output which can only be compared to the precise validation set in terms of hierarchical measures.
    The results are presented in \cref{tbl:limitsofselfsupervisionaccuracy}.
    We include a variation where even the noisy label is withheld from the method, such that all labels must be extrapolated from the root (effectively unsupervised learning).
    Including the noisy label produces an improvement in accuracy of up to $13.9$ pp and is always beneficial.
    
    \textbf{Even in the worst case (iv), more than half of the labels that are extrapolated as far as possible are correct.}
    This is a strong result considering the $555$ classes and only $4.8\%$ precise labels in the training set.

    \subsection{Non-Adaptive Methods}\label{sec:expnonadaptive}
    The study in \cref{sec:limitsofselfsupervision} shows that most cases, except possibly (ii), have a strong
    potential for improvement by using extrapolated predictions as pseudo-labels.
    However, a high hierarchical F1 score in one step does not necessarily generalize to high accuracy when using self-supervision
    continuously during training.
    On the one hand, the aforementioned feedback loop could negatively affect training by overfitting and unbalancing the training data.
    On the other hand, the quality of extrapolations might improve over time as the model learns from correct pseudo-labels.

    We first apply the non-adaptive methods described in \cref{sec:nonadaptivemethods}
    to CHILLAX on NABirds.
    Ground truth labels are replaced with pseudo-labels at all times during training.
    The resulting model is then evaluated in terms of accuracy on the NABirds validation set.
    We repeat the experiment six times per individual setting and include a CHILLAX baseline for comparison.

    \begin{table*}
        \caption{Accuracy (\%) on NABirds validation set.
        Comparison between non-adaptive self-supervised methods on noisy training data.
        Baseline \textnormal{(i)} result $81.63 \pm 0.12$.}
        \label{tbl:nonadaptivemethods}
        \centering
        \begin{tabular}{llllll}
            \toprule
            Method / Noise &                     (ii) &             (iii) &              (iv) &               (v) \\
            \midrule
            CHILLAX Baseline       &    62.66 $\pm$ 0.82 &  49.04 $\pm$ 1.04 &  43.18 $\pm$ 0.20 &  70.91 $\pm$ 0.34 \\
            Leaf Node              &    63.05 $\pm$ 1.37 &  49.36 $\pm$ 0.48 &  43.49 $\pm$ 0.20 &  70.94 $\pm$ 0.42 \\
            \underline{k Steps Down}\\
            $1$                    &    61.78 $\pm$ 0.27 &  33.11 $\pm$ 0.87 &  23.49 $\pm$ 0.60 &  65.44 $\pm$ 0.83 \\
            $1$, conf. $\geq 0.8$  &    61.75 $\pm$ 0.69 &  48.09 $\pm$ 0.75 &  40.85 $\pm$ 1.26 &  71.67 $\pm$ 0.23 \\
            $1$, conf. $\geq  0.9$ &    63.13 $\pm$ 0.70 &  49.98 $\pm$ 0.55 &  41.54 $\pm$ 1.37 &  71.75 $\pm$ 0.26 \\
            $2$                    &    61.31 $\pm$ 0.68 &  14.53 $\pm$ 0.82 &  12.12 $\pm$ 0.96 &  59.58 $\pm$ 0.60 \\
            $2$, conf. $\geq  0.8$ &    62.07 $\pm$ 0.43 &  48.31 $\pm$ 0.84 &  37.21 $\pm$ 0.72 &  71.74 $\pm$ 0.81 \\
            $2$, conf. $\geq  0.9$ &    62.52 $\pm$ 0.68 &  \textbf{50.68} $\pm$ 0.44 &  41.78 $\pm$ 0.47 &  71.54 $\pm$ 0.33 \\
            
            \underline{Threshold}\\
            $0.55  $                 &  61.48 $\pm$ 0.36 &  26.56 $\pm$ 0.94 &  22.68 $\pm$ 0.29 &  65.32 $\pm$ 0.53 \\
            $0.8   $                 &  61.73 $\pm$ 0.54 &  39.80 $\pm$ 0.97 &  31.15 $\pm$ 1.65 &  69.86 $\pm$ 1.23 \\
            $0.85  $                 &  61.93 $\pm$ 0.25 &  43.35 $\pm$ 0.72 &  34.60 $\pm$ 0.73 &  70.59 $\pm$ 0.17 \\
            $0.9   $                 &  62.34 $\pm$ 0.33 &  46.74 $\pm$ 1.27 &  38.03 $\pm$ 0.78 &  \textbf{71.77} $\pm$ 0.00 \\
            $0.95  $                 &  62.75 $\pm$ 0.21 &  48.66 $\pm$ 1.03 &  42.11 $\pm$ 1.48 &  71.47 $\pm$ 0.44 \\
            $0.97  $                 &  63.00 $\pm$ 0.58 &  50.09 $\pm$ 0.26 &  43.20 $\pm$ 0.57 &  71.40 $\pm$ 0.25 \\
            $0.99  $                 &  63.51 $\pm$ 0.52 &  49.37 $\pm$ 0.28 &  \textbf{44.02} $\pm$ 0.12 &  71.14 $\pm$ 0.15 \\
            $0.992 $                 &  63.02 $\pm$ 0.57 &  48.88 $\pm$ 0.65 &  43.78 $\pm$ 0.33 &  71.21 $\pm$ 0.51 \\
            $0.994 $                 &  \textbf{63.54} $\pm$ 0.51 &  49.23 $\pm$ 0.55 &  43.61 $\pm$ 0.94 &  70.99 $\pm$ 0.27 \\
            $0.996 $                 &  63.11 $\pm$ 0.60 &  49.37 $\pm$ 0.61 &  43.69 $\pm$ 1.00 &  71.08 $\pm$ 0.45 \\
            $0.998 $                 &  63.10 $\pm$ 0.83 &  49.16 $\pm$ 0.63 &  43.81 $\pm$ 0.25 &  70.96 $\pm$ 0.22 \\
            $0.999 $                 &  62.78 $\pm$ 0.48 &  49.56 $\pm$ 1.04 &  42.92 $\pm$ 0.68 &  71.37 $\pm$ 0.39 \\
            \bottomrule
            \end{tabular}
            
    \end{table*}

    The results are shown in \cref{tbl:nonadaptivemethods}, where we first observe the threshold-based extrapolation method.
    It is very sensitive to the confidence threshold and requires substantial fine-tuning.
    The optimal threshold strongly depends on the noise model.
    For example, the Poisson noise (iv) only has a working range of thresholds from $0.97$ to $0.998$
    where it matches or outperforms the baseline, with the optimum at $0.99$.
    
    \textbf{Overall, improvements \wrt{} the baseline range from 0.84 pp to 1.05 pp.}

    The \enquote{leaf node} and \enquote{steps down} methods are somewhat competitive, specifically
    for the geometric noise model (iii).
    Going $k$ steps down the hierarchy is only beneficial when combined with
    a confidence threshold. However, this combination suffers from the aforementioned fine-tuning problem.
    
    \textbf{Always selecting the most confident leaf node leads to a small improvement in all cases 
    and requires no tuning.}

    \subsection{Adaptive Self-Supervision}\label{sec:expadaptive}
    This experiment compares our two proposed adaptive methods of limiting the increase in
    IC from extrapolation source to target (see \cref{sec:adaptivemethods} for details).
    The first proposed adaptive method \enquote{adaptive threshold} changes a
    confidence threshold dynamically to achieve a given expected IC gain.
    Our experiment uses expected gains of $0.025, \ldots{}, 0.1$, where our choice of IC is naturally bounded between $0$ and $1$.
    The second method \enquote{IC range} uses a fixed range of allowed IC differences.
    We use the ranges $[0,0.2],[0.1,0.3],\ldots{},[0.4,0.6]$ and a minimum confidence of $0.55$ to reject spurious predictions.
    We perform six training repetitions for each combination of method and noise model.

    \begin{table*}
        \caption{Accuracy (\%) on NABirds validation set.
        Comparison between adaptive self-supervised methods on noisy training data.
        Baseline \textnormal{(i)} result $81.63 \pm 0.12$.
        \textbf{Best adaptive result}, \underline{Best overall result}}
        \label{tbl:adaptivemethods}
        \centering
        \begin{tabular}{lllll}
            \toprule
            Method / Noise &              (ii) &             (iii) &              (iv) &               (v) \\
            \midrule
            CHILLAX Baseline       &  62.66 $\pm$ 0.82 &  49.04 $\pm$ 1.04 &  43.18 $\pm$ 0.20 &  70.91 $\pm$ 0.34 \\
            Best non-adaptive      &  \underline{63.54 $\pm$ 0.51} &  \underline{50.68 $\pm$ 0.44} &  \underline{44.02 $\pm$ 0.12} &  71.77 $\pm$ 0.00 \\
            \underline{Adaptive Threshold}\\
            $0.025$                &  61.80 $\pm$ 0.69 &  49.07 $\pm$ 0.93 &  \textbf{43.35 $\pm$ 0.82} &  71.58 $\pm$ 0.40 \\
            $0.0375$               &  61.43 $\pm$ 0.51 &  \textbf{49.75 $\pm$ 1.13} &  43.11 $\pm$ 1.49 &  72.00 $\pm$ 0.43 \\
            $0.05$                 &  61.80 $\pm$ 0.44 &  49.52 $\pm$ 0.90 &  42.92 $\pm$ 0.42 &  \underline{\textbf{72.10 $\pm$ 0.31}} \\
            $0.0625$               &  61.87 $\pm$ 0.91 &  49.57 $\pm$ 1.55 &  42.30 $\pm$ 0.35 &  71.15 $\pm$ 0.92 \\
            $0.075$                &  62.20 $\pm$ 0.76 &  49.97 $\pm$ 0.59 &  41.71 $\pm$ 1.13 &  71.37 $\pm$ 0.40 \\
            $0.1$                  &  61.79 $\pm$ 0.30 &  48.57 $\pm$ 0.61 &  39.72 $\pm$ 0.96 &  70.77 $\pm$ 0.34 \\
            \underline{Fixed Range}\\
            $[0,0.2]$              &  62.11 $\pm$ 0.44 &  46.42 $\pm$ 0.71 &  40.50 $\pm$ 0.50 &  68.39 $\pm$ 0.47 \\
            $[0.1,0.3]$            &  61.78 $\pm$ 0.42 &  29.77 $\pm$ 0.98 &  26.07 $\pm$ 0.67 &  64.31 $\pm$ 1.05 \\
            $[0.2,0.4]$            &  \textbf{63.43 $\pm$ 0.42} &  36.00 $\pm$ 1.29 &  30.64 $\pm$ 0.86 &  68.61 $\pm$ 0.22 \\
            $[0.3,0.5]$            &  63.23 $\pm$ 0.10 &  35.00 $\pm$ 0.46 &  31.45 $\pm$ 0.60 &  68.55 $\pm$ 0.94 \\
            $[0.4,0.6]$            &  62.96 $\pm$ 0.98 &  33.24 $\pm$ 0.65 &  27.31 $\pm$ 1.41 &  68.46 $\pm$ 0.33 \\
            \bottomrule       
            \end{tabular}
    \end{table*}

    \Cref{tbl:adaptivemethods} compares the results of both methods.
    Our adaptive threshold method performs better than the fixed range approach in the noisier settings (iii)-(v),
    even outperforming the fine-tuned non-adaptive fixed threshold method on setting (v) with in improvement in accuracy of $1.19$ pp.
    
    \textbf{Furthermore, the parameter of our adaptive threshold method is much less sensitive to changes than the fixed threshold
    as evidenced by the large effective range.}

    The fixed IC gain range setup only works well for noise model (ii), which is the immediate parent relabeling scenario from \cite{deng_large-scale_2014}.
    This result is expected, because this noise model leaves only two possibilities for IC gain.
    There can either be no gain at all, or the fixed amount when moving from the second to last level in the hierarchy to the last level.
    As such, the model fits the assumption of a fixed IC gain range perfectly.
    The other noise models lead to partly catastrophic results when the fixed range effectivly prohibits any extrapolation.
    However, if the noise distribution is known before training, it could be argued that setting a correct range is more straightforward.

    \paragraph{}%
    Overall, we observe that a fixed threshold performs best, but only after significant fine-tuning.
    \textbf{Our \enquote{adaptive threshold} method is less sensitive to changes in its parameters,
    and performs slightly better than the parameter-free \enquote{leaf node} approach,
    which is why we recommend it as a drop-in replacement for fully supervised CHILLAX.}
    
    \section{CONCLUSION}
    Learning from imprecise data is proposed in \cite{Brust2020ELC} as a way of maximizing training data utilization.
    However, their method CHILLAX does not utilize \emph{all} training data as it ignores examples labeled at the root.
    This is a consequence of the closed-world assumption made by the underlying classifier.
    To avoid such meaningless labels, we propose to use CHILLAX's label extrapolation not just at test time, but during training to generate pseudo-labels.
    This increases the precision of examples labeled not only at the root, but also at inner nodes of the class hierarchy.

    To implement this self-supervised learning scheme, we describe several possible strategies of deciding which candidate pseudo-labels are reliable enough for training.
    These strategies employ heuristic, structural and statistical criteria.
    Our experiments show that an increase in accuracy of around one percent point can be expected by simply using one of our self-supervised strategies on top of CHILLAX.
    This improvement comes without any requirement of fine-tuning unrelated parameters or undue computational efforts.

    \paragraph{Future Work}
    In the future, these methods could also be applied to semi-supervised learning tasks in general, \eg{}, by assigning a root label to the unlabeled images as long as a closed-world scenario can be assumed.
    Furthermore, the individual heuristics could be combined into a meta-heuristic.
    In contrast, relaxing the closed-world assumption is another important research direction.
    Asking a hierarchical classifier for its confidence in the root node is a first step towards open-set models from a semantic perspective, as long as the predicted confidence has a reasonable basis.
    A fixed hierarchy is a further limiting assumption, which could be relaxed, \eg{}, in a lifelong learning setting.

    The research on semantically imprecise data in general could be expanded to domains beyond natural images.
    For example, we expect source code to have a stronger feature-semantic correspondence, which is crucial for the hierarchical classifier.
    In particular, human-made hierarchies such as the Common Weakness Enumeration (CWE, \cite{mitrecwe}) explicitly consider certain features of program code to determine categories.
    And even in the visual domain, there are efforts to construct more visual-feature-oriented hierarchies, \eg{}, accompanying WikiChurches \cite{barz2021wikichurches}.

    \emph{Acknowledgments:} The computational experiments were performed on resources of Friedrich
    Schiller University Jena supported in part by DFG grants INST 275/334-1 FUGG and INST 275/363-1 FUGG. 


    \bibliographystyle{apalike}
    {\small
    \bibliography{paper}}

\begin{thebibliography}{}

\bibitem[Abassi and Boukhris, 2020]{abassi_imprecise_2020}
Abassi, L. and Boukhris, I. (2020).
\newblock Imprecise label aggregation approach under the belief function
  theory.
\newblock In Abraham, A., Cherukuri, A.~K., Melin, P., and Gandhi, N., editors,
  {\em Intelligent Systems Design and Applications}, pages 607--616. Springer
  International Publishing.

\bibitem[Ambroise et~al., 2001]{ambroise_learning_2001}
Ambroise, C., Denœux, T., and Govaert, G. (2001).
\newblock Learning from an imprecise teacher: probabilistic and evidential
  approaches.
\newblock {\em Applied Stochastic Models and Data Analysis}, page~6.

\bibitem[Barz and Denzler, 2021]{barz2021wikichurches}
Barz, B. and Denzler, J. (2021).
\newblock Wikichurches: A fine-grained dataset of architectural styles with
  real-world challenges.
\newblock In {\em Neural Information Processing Systems (NeurIPS)}.

\bibitem[Brust et~al., 2020]{Brust2020ELC}
Brust, C.-A., Barz, B., and Denzler, J. (2020).
\newblock Making every label count: Handling semantic imprecision by
  integrating domain knowledge.
\newblock In {\em International Conference on Pattern Recognition (ICPR)}.

\bibitem[Brust and Denzler, 2019]{brust_integrating_2018}
Brust, C.-A. and Denzler, J. (2019).
\newblock Integrating domain knowledge: using hierarchies to improve deep
  classifiers.
\newblock In {\em Asian Conference on Pattern Recognition (ACPR)}.

\bibitem[Deng et~al., 2014]{deng_large-scale_2014}
Deng, J., Ding, N., Jia, Y., Frome, A., Murphy, K., Bengio, S., Li, Y., Neven,
  H., and Adam, H. (2014).
\newblock Large-scale object classification using label relation graphs.
\newblock In {\em European Conference on Computer Vision ({ECCV})}.

\bibitem[Deng et~al., 2012]{deng2012hedging}
Deng, J., Krause, J., Berg, A.~C., and Fei-Fei, L. (2012).
\newblock Hedging your bets: Optimizing accuracy-specificity trade-offs in
  large scale visual recognition.
\newblock In {\em Computer Vision and Pattern Recognition (CVPR)}.

\bibitem[Harispe et~al., 2015]{harispe_semantic_2015}
Harispe, S., Ranwez, S., Janaqi, S., and Montmain, J. (2015).
\newblock Semantic similarity from natural language and ontology analysis.
\newblock {\em Synthesis Lectures on Human Language Technologies}, 8(1):1--254.

\bibitem[Kolesnikov et~al., 2019]{Kolesnikov_2019_CVPR}
Kolesnikov, A., Zhai, X., and Beyer, L. (2019).
\newblock Revisiting self-supervised visual representation learning.
\newblock In {\em Proceedings of the IEEE/CVF Conference on Computer Vision and
  Pattern Recognition (CVPR)}.

\bibitem[Lee, 2013]{lee_pseudo-label_2013}
Lee, D.-H. (2013).
\newblock Pseudo-label : The simple and efficient semi-supervised learning
  method for deep neural networks.
\newblock In {\em International Conference on Machine Learning Workshops
  (ICML-WS)}, page~6.

\bibitem[McAuley et~al., 2013]{mcauley2013optimization}
McAuley, J.~J., Ramisa, A., and Caetano, T.~S. (2013).
\newblock Optimization of robust loss functions for weakly-labeled image
  taxonomies.
\newblock {\em International Journal of Computer Vision (IJCV)},
  104(3):343--361.

\bibitem[Silla and Freitas, 1 01]{silla_survey_2011}
Silla, C.~N. and Freitas, A.~A. (2011-01).
\newblock A survey of hierarchical classification across different application
  domains.
\newblock {\em Data Mining and Knowledge Discovery}, 22(1):31--72.

\bibitem[Sohn et~al., 2020]{sohn_fixmatch_2020}
Sohn, K., Berthelot, D., Li, C.-L., Zhang, Z., Carlini, N., Cubuk, E.~D.,
  Kurakin, A., Zhang, H., and Raffel, C. (2020).
\newblock {FixMatch}: Simplifying semi-supervised learning with consistency and
  confidence.

\bibitem[{The MITRE Corporation}, 2021]{mitrecwe}
{The MITRE Corporation} (2021).
\newblock {Common Weakness Enumeration (CWE)}.

\bibitem[Van~Horn et~al., 2015]{van_horn_building_2015}
Van~Horn, G., Branson, S., Farrell, R., Haber, S., Barry, J., Ipeirotis, P.,
  Perona, P., and Belongie, S. (2015).
\newblock Building a bird recognition app and large scale dataset with citizen
  scientists: {The} fine print in fine-grained dataset collection.
\newblock In {\em Proceedings of the {IEEE} {Conference} on {Computer} {Vision}
  and {Pattern} {Recognition}}, pages 595--604.

\bibitem[Wang et~al., 2016]{wang_cost-effective_2016}
Wang, K., Zhang, D., Li, Y., Zhang, R., and Lin, L. (2016).
\newblock Cost-effective active learning for deep image classification.
\newblock {\em Circuits and Systems for Video Technology ({CSVT})},
  {PP}(99):1--1.

\bibitem[Xie et~al., 2016]{xie2016disturblabel}
Xie, L., Wang, J., Wei, Z., Wang, M., and Tian, Q. (2016).
\newblock Disturblabel: Regularizing cnn on the loss layer.
\newblock In {\em Computer Vision and Pattern Recognition (CVPR)}.

\bibitem[Zhai et~al., 2019]{Zhai_2019_ICCV}
Zhai, X., Oliver, A., Kolesnikov, A., and Beyer, L. (2019).
\newblock S4l: Self-supervised semi-supervised learning.
\newblock In {\em Proceedings of the IEEE/CVF International Conference on
  Computer Vision (ICCV)}.

\end{thebibliography}
\end{document}